\documentclass[a4paper]{styles/svproc}
\usepackage{url}
\usepackage{amsmath}
\usepackage{siunitx}
\usepackage{amssymb}
\usepackage{hyperref}
\usepackage{float}

\usepackage{listings}
\usepackage{algorithm}
\usepackage{algpseudocode}

\usepackage[export]{adjustbox}
\usepackage{graphicx}
\usepackage[square,numbers]{natbib}
\usepackage{caption}
\usepackage{subcaption}
\usepackage{wrapfig}
\usepackage{tikz}

\usepackage{xspace}

\def\R{\mathbb{R}}
\def\D{D}
\def\X{X}
\def\Y{Y}

\def\Xf{\X_f}
\def\P{P}
\def\Pf{\P_f}
\def\costfunctional{J}
\def\optimizer{\Phi_\costfunctional}

\def\x{x}

\def\xfk{\x_{\Fk}}
\def\xkk{\x_{k+1}}
\def\planningproblem{\left(\Xf,\x_I,\x_G\right)}

\def\simeq{\sim_{\optimizer}}
\def\simeqproj{\sim_{\{\optimizer,\pi\}}}

\def\timebudget{t_b}

\def\v{v}
\def\vprime{{\v}^{\prime}}
\def\p{p}
\def\pprime{{\p}^{\prime}}
\def\pstar{{\p}^{\star}}
\def\q{q}
\def\qk{{\q}_k}
\def\qprime{{\q}^{\prime}}

\def\tree{T}

\def\fiber{F}
\def\F{\fiber}
\def\Fk{{\fiber}_k}
\def\Xk{{\X}_k}
\def\Xkk{{\X}_{k+1}}

\def\fiberBundleSequence{\ensuremath{\X_K \overset{\pi_K}{\longrightarrow} \X_{K-1}
\overset{\pi_{K-1}}{\longrightarrow} \cdots
\overset{\pi_1}{\longrightarrow}
\X_0}}

\def\G{\ensuremath{\mathbf{G}}}
\def\S{\ensuremath{\mathbf{S}}}
\def\Gk{\G_k}
\def\Gkk{\G_{k+1}}
\def\Sk{\S_k}
\def\Skk{\S_{k+1}}

\def\deltaS{\delta_S}
\def\tree{T}

\def\xr{x_{\text{rand}}}

\makeatletter
\newcommand{\toprule}{\hrule height.8pt depth0pt \kern2pt} 
\newcommand{\midrule}{\kern2pt\hrule\kern2pt} 
\newcommand{\bottomrule}{\kern2pt\hrule\relax}
\newcommand{\algcaption}[2][]{%
  \refstepcounter{algorithm}%
  \toprule
  \textbf{{\raggedright\fname@algorithm~\thealgorithm}}\ #2\par 
  \midrule
}
\makeatother

\begin{document}
\mainmatter

\title{Visualizing Local Minima in Multi-Robot Motion Planning using Multilevel Morse Theory}

\titlerunning{Visualizing Local Minima}

\author{Andreas Orthey\inst{1} \and Marc Toussaint\inst{1,2}}
\authorrunning{Andreas Orthey and Marc Toussaint}
\institute{Max Planck Institute for Intelligent Systems, Stuttgart, Germany\\
\email{aorthey@is.mpg.de}
\and
Technical University of Berlin, Germany
}

\maketitle 

\begin{abstract}

Multi-robot motion planning problems often have many local minima. It is
  essential to visualize those local minima such that we can better understand,
  debug and interact with multi-robot systems. Towards this goal, we present the
  multi-robot motion explorer, an algorithm which extends previous results on
  multilevel Morse theory by introducing a component-based framework, where
  we reduce multi-robot configuration spaces by reducing each robots component space using fiber bundles. Our algorithm exploits this component structure to search for and visualize local minima. A user of the algorithm can specify a multilevel
  abstraction and an optimization algorithm. We use this information to incrementally build a
  local minima tree for a given problem. We
  demonstrate this algorithm on several multi-robot systems of up to 20 degrees
  of freedom.

\keywords{Multi-robot motion planning, Morse Theory, Fiber bundles, Visualization of Local Minima}
\end{abstract}

\section{Introduction}



Coordinating multiple robots is essential to automate surveillance of climate
changes, to collaborate on construction sites and to route autonomous vehicles.
Problems of coordinating multiple robots often involve many local minima. A
local minimum is a solution path to a multi-robot planning problem, such that we
cannot (locally) improve upon the path, i.e. the path is
\emph{invariant under optimization of a cost functional} \cite{orthey_2020}. Existing algorithms,
however, usually do not compute or visualize those local minima and often search
for only a single solution \cite{lavalle_2006}.

To solve multi-robot motion planning problems, we argue it to be essential to
visualize local minima. By visualizing local minima, we can obtain conceptual
understanding about the underlying topological complexity \cite{smale_1987},
extract symbolic representations \cite{toussaint_2017} and analyse the
convergence of optimization algorithms \cite{zucker_2013}. By visualizing local
minima, we allow interaction by non-expert users, to either guide or prevent
motions \cite{orthey_2020}. By visualizing local minima, we can create
high-level options \cite{pall_2018}, usable to make high-level decisions
\cite{sontges_2017} or perform rapid re-planning \cite{yang_2010}.


Visualizing local minima is therefore an important requirement for multi-robot
motion planning. To enumerate local minima, we develop a new algorithm we call
the \emph{multi-robot motion explorer}. The explorer extends previous works on
single-robot motion planning \cite{orthey_2020}. In particular, we evoke Morse
theory \cite{morse_1934} to enumerate local minima. To each local minimum we
assign an equivalence class of paths converging, under optimization, to the same
local minimum. Using those equivalence classes, we use a predefined multilevel
representation, described by fiber bundles \cite{orthey_2019}, to organize local
minima into a local-minima tree. Eventually, non-expert users can interact with
this tree by navigating through it similar to navigating through a unix
filesystem.

Our contributions are

\begin{enumerate}

    \item An algorithm, we call the multi-robot motion explorer, using a
      component-based framework to incrementally build local minima trees for
      multi-robot planning problems

    \item Demonstration of the multi-robot motion explorer on six multi-robot
      planning problems of up to $20$ degrees of freedom

\end{enumerate}


\section{Related Work}

To visualize local minima, we need to solve two associated problems. First, we need to find a representation of the configuration space. Second, we need to utilize this representation to extract local minima.

\subsection{Multi-Robot Motion Planning}

To represent a multi-robot motion planning problem, we can consider the robots as one generalized robot under robot-robot collision constraints in a composite configuration space \cite{lavalle_2006}. In general, we can use such an approach only for a few low-dimensional robots, mainly because the problem itself is NP-hard \cite{hopcroft_1984}. Since the problem is NP-hard, it becomes necessary to reduce the composite configuration space. Depending on the type of robots, we can group reduction methods into two classes.

First, we have reductions for the case where all robots are equivalent (homogeneous). For homogeneous multi-robot systems, we can project all start and goal configurations into the configuration space of the first robot and find a graph connecting all configurations. To coordinate the motion of the robots along that graph we need to solve the pebbles-on-a-graph problem, which we can solve efficiently \cite{kornhauser_1984}, either by converting it to an integer linear program \cite{yu_2016}, by partitioning the graph into regions densely or sparsely connected \cite{ryan_2010} or by utilizing simple push and swap strategies \cite{luna_2011}. By utilizing solvers for pebbles-on-a-graph, we can for example create a larger framework to compute motions for swarms of drones \cite{hoenig_2018}.

Second, we have reductions for the case where robots are not equivalent (non-homogeneous). For non-homogeneous multi-robot systems, we can first compute graphs on each component configuration space and then merge them into a graph on the composite configuration space \cite{lavalle_2006}. To merge graphs, we can either use path coordination or graph coordination.

In path coordination \cite{simeon_2002}, we compute paths for each robot separately. We then coordinate the execution of those paths, either by searching over the space of reparameterizations \cite{simeon_2002} or by prioritizing the robots \cite{erdmann_1987}.

In graph coordination, we compute graphs for each robot separately. We then combine the graphs into an (implicit) composite configuration space graph \cite{svestka_1998}. To compute this graph, we can use two methods. 

First, we can create the \emph{tensor product} of graphs, whereby all edges are combined \cite{solovey_2016}. To explore the tensor product, we can either use a random search utilizing a direction-oracle \cite{solovey_2016} or we can execute shortest paths optimistically until conflicts arise. When conflicts arise, we can expand locally the dimensionality to resolve conflicts \cite{wagner_2015}. We can combine both methods using prioritization of robots, for example by analyzing possible start and goal conflicts \cite{vandenberg_2009} or by the number of topologically varying paths a robot can execute \cite{wu_2019}.

Second, we can create the \emph{cartesian product} of graphs, whereby only edges are used where at most one robot moves. We can think of the cartesian product as an approximation to the tensor product, in the sense that every multi-robot path can be arbitrarily close approximated by a path where at most one robot moves at a time \cite{svestka_1998}. However, if the underlying graphs are not dense enough, we can miss valid paths and thereby sacrifice completeness.

It is important to note that non-homogenous and homogenous robots are not mutually exclusive, but we can often further simplify non-homogenous robot problems by decomposing them into problems of groups of homogenous robots \cite{solovey_2014}. 

\subsection{Multi-Path Multi-Robot Motion Planning}

From a given representation of the composite configuration space, we like to extract local minima. Local minima can often be defined as representative paths of equivalence classes, where we define an equivalence relation on the path space for the purpose of grouping paths. Grouping paths can be done using different approaches, of which we discuss three fundamental ones. 

First, we can group paths topologically \cite{munkres_1974}. In a topologically grouping, we use the notion of homotopy to define two paths to be equivalent if they can be continuously deformed into each other. To find paths which differ homotopically, we can compute a simplicial complex of the configuration space and extract paths \cite{pokorny_2016} or by computing an H-score determining the number of times a path crosses subsets of the configuration space \cite{bhattacharya_2018}. 

Second, we can group paths based on braid patterns \cite{artin_1947}. In a braid pattern grouping, we define two paths to be equivalent if pairwise robot crossings are equivalent. By ignoring the type of crossing, we obtain a permutation group of robots. Using this permutation group, we can compute representative paths of varying braid pattern \cite{mavrogiannis_2016}.
We can alternatively find paths of varying braid pattern 
by planning a minimal-cost path constrained to a pattern \cite{mavrogiannis_2018} or by following a braid pattern controller, for example with safety separations \cite{diazmercado_2017}. 

Third, we can group paths based on Morse theory \cite{morse_1934}. In morse theory, we define two paths to be equivalent if they converge, under optimization, to the same local minimum \cite{orthey_2020}. This differs from braid theory and topology (1) by being finer in the sense that two equivalent paths under braid theory or topology can converge to two different minima, (2) by being defined relative to an optimizer and (3) by being often faster to compute in higher dimensions \cite{jaillet_2008}. In previous work we used Morse theory in single-robot motion planning \cite{orthey_2020}. In this work, we generalize this approach to multi-robot motion planning.

\section{Foundations}

Let $r_1,\ldots, r_M$ be $M$ robots with associated \emph{component}
configuration spaces $\Y_1,\ldots, \Y_M$ of dimensionality $n_1,\ldots, n_M$. We
define the (composite) configuration space $\X = \Y_1 \times \cdots \times \Y_M$
of dimensionality $n = n_1 + \cdots + n_M$. The space of constraint-free
configurations is denoted as $\Xf$. The motion planning problem
$\planningproblem$ asks us to find a path from a start configuration $x_I \in
\Xf$ to a goal configuration $x_G \in \Xf$. 

The space of solutions to the motion planning problem is given by the associated
path space. The path space $\Pf$ is the space of all continuous paths $p: I
\rightarrow \Xf$ on $\Xf$ starting at $x_I$ and ending at $x_G$. To analyse
$\Pf$, we enumerate local minima using Morse theory and use multilevel abstractions represented by fiber bundles to
organize local minima into a local minima tree.

\subsection{Morse Theory\label{sec:morsetheory}}

We utilize Morse theory \cite{morse_1934} to identify local minima. A local
minimum is an invariant of the path space $\Pf$ under optimization of a cost
functional. A cost functional $\costfunctional$ maps a path $p \in \Pf$ to a
real number $\R$ as 

\begin{equation}\label{eq:costfunctional}
    \costfunctional[p] = \int_0^1 L(x,p(x),p'(x)) dx 
\end{equation}

whereby $L$ is a loss term. To solve Eq. \ref{eq:costfunctional}, we can take
one of two views.  In the first view, we interpret Eq. \eqref{eq:costfunctional}
as a problem of optimal control or calculus of variation in the small
\cite{gelfand_2000}, where we like to find \emph{one} global minimal solution.
In the second view, however, we interpret Eq. \eqref{eq:costfunctional} as a
problem of Morse theory or calculus of variation in the large \cite{morse_1934},
where we like to find \emph{all} local minimal solutions. 

In this paper, we adopt the Morse theoretic view to enumerate local minima. We
define a local minimum as an invariant of Eq. \eqref{eq:costfunctional} under
optimization. To optimize, we use a local optimizer $\optimizer: \Pf \rightarrow
\Pf$ which we assume to be given. We require $\optimizer$ to be different from an
identity mapping, but make no other assumption about its behavior.

Following Morse theory, we interpret the optimizer $\optimizer$ as an
equivalence relation\footnote{Recall that an equivalence relation $\sim$ on a
path space $\P$ is a binary relation such that for any paths $a,b,c \in \P$ we
have $a \sim a$ (Reflexive), if $a \sim b$ then $b \sim a$ (Symmetric) and if $a
\sim b$ and $b \sim c$ then $a \sim c$ (Transitive) \cite{munkres_1974}.} on the
path space \cite{orthey_2020}. In particular, given two paths $\p, \pprime \in
\Pf$ we define them to be equivalent, written as $\p \simeq \pprime$, if
$\optimizer(\p) = \optimizer(\pprime)$.  We then take the quotient $Q =
\Pf/\simeq$ which represents equivalence classes of paths under optimization.
Each equivalence class will be \emph{represented} by one path invariant under
optimization, i.e. a \emph{local minimum} $\pstar$ for which we have
$\optimizer(\pstar) = \pstar$. Under this representation, we associate to every
$\Xf$ its local-minima space $Q$ containing all local-minima of $\Pf$. 

\subsection{Multilevel Abstractions using Fiber Bundles}

Finding and interpreting local minima is often too difficult in high-dimensional
configuration spaces. To simplify those spaces, we use multiple levels of
abstractions which we model using the language of fiber bundles
\cite{steenrod_1951}. Fiber bundles are sets of admissible lower-dimensional
projections we use to decrease planning time \cite{orthey_2019} and to
group local minima into more meaningful classes \cite{orthey_2020}.

\begin{wrapfigure}{r}{0.45\linewidth}

    \centering

\begin{tikzpicture}
\def\height{2}
\def\radius{1.25}
\def\heightEllipse{0.5}
\def\distBase{1.8}
\def\pathLength{1.5}
\def\vertexStart{-0.5*\pathLength}
\def\pathStretch{2*3.1415/\pathLength}

\def\xposFiber{-2.2*\radius}
\def\xposPath{0*\radius}
\def\xposGraph{2.2*\radius}

\tikzset{
    classical/.style={thin,double,->,shorten >=4pt,shorten <=4pt,>=stealth}
}
\newcommand\drawCylinder[1]{
\draw (#1,\height) ellipse ({\radius} and \heightEllipse);
\draw (#1-\radius,0) -- (#1-\radius,\height);
\draw (#1+\radius,0) -- (#1+\radius,\height);
\draw (#1-\radius,0) arc (180:360:{\radius} and \heightEllipse);
\draw [dashed] (#1-\radius,0) arc (180:360:{\radius} and -\heightEllipse);
\draw (#1,-\distBase) ellipse ({\radius} and \heightEllipse);
\draw[classical] (#1,0-\heightEllipse) -- (#1,-\distBase+\heightEllipse);
}

\newcommand\drawFiber[1]{
\def\pointPosX{#1+0.5*\radius}
\draw[line width=0.3mm,black](\pointPosX,0) -- (\pointPosX, \height);
\node[draw,fill,circle,inner sep=0pt,minimum size=0.3mm] at (\pointPosX, 0-\distBase){};
\draw node[left] at (\pointPosX, 0-\distBase) {$x_B$};
\draw node[left] at (\pointPosX, 0.5*\height) {$\pi^{-1}(x_B)$};
}

\drawCylinder{\xposFiber}
\drawFiber{\xposFiber}

\end{tikzpicture}
    \caption{Fiber bundle $\D^2 \times \R^1 \rightarrow \D^2$\label{fig:fiberbundle}}
    
\end{wrapfigure}
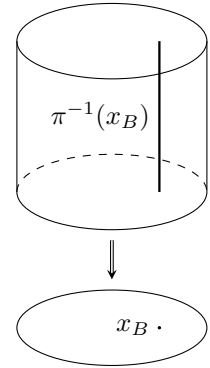

Formally, a fiber bundle is a tuple $(\X, F, B)$ with a projection mapping
\begin{equation}
    \pi: \X \longrightarrow B
\end{equation}

whereby $\X$ is the bundle space and $B$ the base space. Both bundle and base space have associated constraint-free subspaces $\Xf$ and $B_f$.

We impose three restriction on the fiber bundle. First, the preimage $\pi^{-1}(x_B)$ (called the fiber over $x_B$) of an element $x_B \in B$ is required to be isomorphic to the fiber space $F$. Second, the bundle space $X$ needs to be (locally) a product space $B \times F$ \cite{lee_2003}. Third, we require the projection map to be admissible, meaning that the projection of $\Xf$ is a subset of $B_f$ \cite{orthey_2019}. With the last requirement we ensure that no local minima are removed after a projection \cite{orthey_2020}. 

We will often abbreviate a
fiber bundle using the shorthand $\X \overset{\pi}{\longrightarrow} B$. As an example, we visualize in Fig. \ref{fig:fiberbundle} the fiber bundle $\D^2 \times \R^1 \rightarrow \D^2$ whereby $\D^2$ is the 2-d disk. The fiber space in this case is $\R^1$ which is isomorphic to the preimage $\pi^{-1}(x_B)$ of a point $x_B$ in $\D^2$. To get a better understanding of fiber bundles, we like to think of the preimages as equivalence classes of $\X$ which are collapsed during projection onto the (quotient) base space $B$. 

A fiber bundle represents a single level of abstraction. However, we often like
to represent configuration spaces on multiple levels of abstraction. In those cases, we use a
\emph{fiber bundle sequence}. A fiber bundle sequence $\X_K
\overset{\pi_{K-1}}{\longrightarrow} \X_{K-1} \overset{\pi_{K-2}}{\longrightarrow}
\cdots \overset{\pi_1}{\longrightarrow} \X_1$ with $\X_K = \X$ consists of $K$
bundle spaces and $K-1$ projection mappings, whereby each mapping adheres to the aforementioned requirements.

In a motion planning problem, we use fiber bundle sequences to describe simplifications, either by removing links from a robot, shrinking links, removing robots or by nesting simpler robots with less degrees-of-freedom. 

\begin{figure}[t]
    \centering
    \begin{subfigure}[t]{0.34\linewidth}
    \centering
    \includegraphics[width=\linewidth]{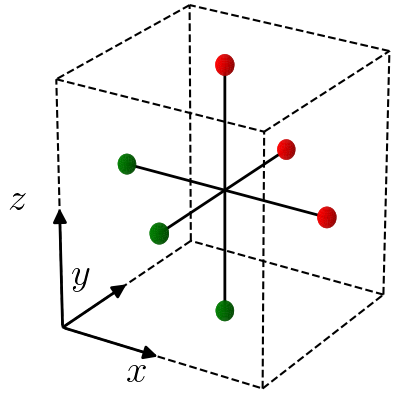}
    \caption{Workspace}
    \label{fig:localminimatree:workspace}
    \end{subfigure}
    \begin{subfigure}[t]{0.34\linewidth}   
    \includegraphics[width=\linewidth]{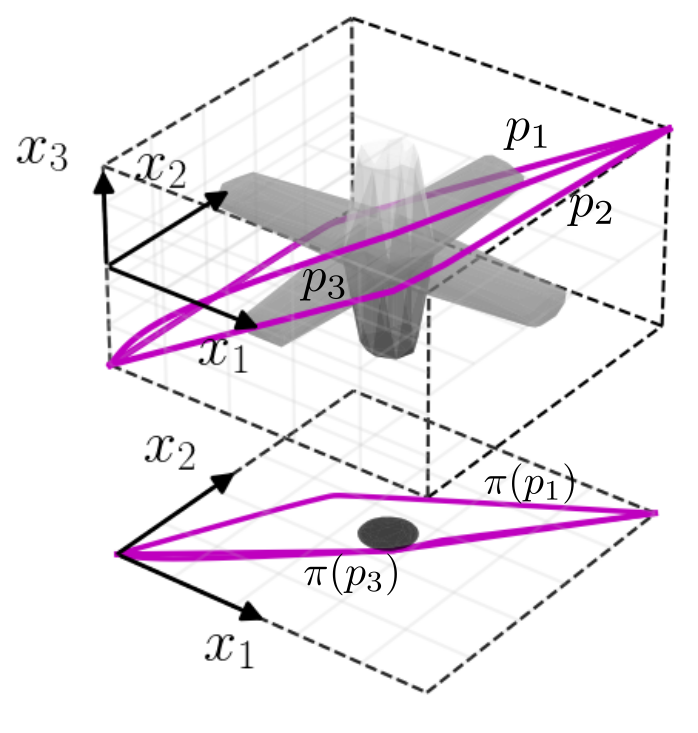}
    \caption{Fiber bundle $\R^3 \rightarrow \R^2$}
    \label{fig:localminimatree:bundlespace}
    \end{subfigure}
    \begin{subfigure}[t]{0.26\linewidth}
    \includegraphics[width=\linewidth]{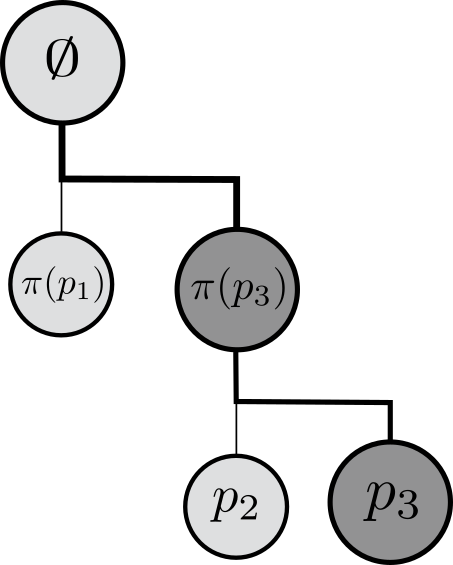}
    \caption{Local-Minima Tree}
    \label{fig:localminimatree:tree}

    \end{subfigure}
    \caption{Local Minima Tree for three spherical robots on line segments. See text for clarification.}
    \label{fig:localminimatree}
\end{figure}

\subsection{Local-Minima Tree for Multilevel Morse theory\label{sec:foundation:localminimatree}}

By combining Morse theory and fiber bundles, we can organize local minima into a
local minima tree. Let $\X_K \overset{\pi_{K-1}}{\longrightarrow} \X_{K-1}
\overset{\pi_{K-2}}{\longrightarrow} \cdots \overset{\pi_1}{\longrightarrow}
\X_1$ be a fiber bundle sequence and let $Q_K$ be the space of local minima
associated to $\X_K=\X$ (Sec. \ref{sec:morsetheory}). 

Using the projection mappings, we can reduce the space $Q_K$ by introducing the
notion of projection-equivalence. Two minima $\q, \qprime \in Q_K$ are said to be
projection-equivalent, written as $\q \simeqproj \qprime$, if
$\optimizer(\pi_K(\q)) = \optimizer(\pi_K(\qprime))$. From this equivalence
relation, we can create the quotient space $Q_{K-1} = Q_K/\simeqproj$. We can
then iteratively apply projection-equivalence using $\pi_{K-1}, \pi_{K-2}, \ldots$ to
obtain a sequence of local-minima spaces $Q_K, Q_{K-1}, \ldots, Q_1$
\cite{orthey_2020}.

Finally, we can organize the local minima spaces into a local-minima tree.  The
local minima tree $\tree = (V, E)$ consists of all elements of the
local-minima spaces as vertices $V$ and of a set of directed edges $E$. An edge
exists between two vertices $\v$ and $\vprime$, if $\vprime$ is an element of $\Xkk$, $\v$ an element of $\Xk$ and if $\v$ is equivalent to
$\vprime$ after projection and subsequent optimization, i.e. $\v =
\optimizer(\pi_k(\vprime))$. We additionally add a root vertex which has a directed edge to every element of $Q_1$.

As an example, we visualize a local minima tree for a 3-dof multi-robot problem in Fig. \ref{fig:localminimatree}. The workspace is depicted in Fig. \ref{fig:localminimatree:workspace}, where we show three spherical robots at a start (green) and a goal configuration (red). The robots are allowed to move exclusively on the line segments between start and goal. The resulting configuration space is a 3d cube and we impose upon it a fiber bundle $\R^3 \overset{\pi}{\longrightarrow} \R^2$, which corresponds to the removal of the third sphere (Fig. \ref{fig:localminimatree:bundlespace}). Inside the cube, we have infeasible regions where the spheres are in collision (grey areas). On the bundle space we visualize three local minima paths (magenta). The two paths $p_2$ and $p_3$ are projection-equivalent when projected onto $\R^2$. This gives a local minima tree, we depict in Fig. \ref{fig:localminimatree:tree}. The tree shows a particular selection of path $p_3$ by a user (dark grey nodes), while we collapse the unselected node $p_1$ for visualization purposes. 
\section{Multi-Robot Motion Explorer}

The multi-robot motion explorer is an algorithm we use to incrementally build a
local minima tree for multi-robot planning problems. A user of the algorithm has
to specify as input a planning problem, a fiber bundle and a
path optimization algorithm. To represent the problem, we grow sequentially and
simultaneously sparse roadmaps on each bundle space. At each iteration of the
algorithm, a user can specify a local minimum to expand. We use this minimum to
grow one of the sparse roadmaps, update the local minima tree and  visualize
the tree to the user. The algorithm differs from previous work
\cite{orthey_2020} by using a component-based fiber bundle method which allows its usage for multi-robot planning problems.

We show the details of the multi-robot motion explorer in Alg.
\ref{alg:explorer}. As input, we are given a planning problem
$\planningproblem$, a prespecified fiber bundle $\fiberBundleSequence$ and an
optimizer $\optimizer$ which minimizes a cost functional $\costfunctional$.
Additional inputs are $N$, the maximum number of returned minima per iteration, $\timebudget$, the time budget for one iteration,
$\deltaS$, the visibility radius used to create a sparse graph, $\epsilon$, a
sampling perturbation parameter and $\rho$ a trade-off between graph sampling
and sampling towards a selected local minimum.

Before starting the algorithm, we initialize each bundle space $\Xk$ by
associating with it a dense graph $\Gk$, a sparse graph $\Sk$ and a fiber space
$\Fk = \Xkk / \Xk$. Extending previous work, we use a new component-based fiber
space method computing $\F^k_m \leftarrow \Y^{k+1}_m / \Y^k_m$
for each robot $m$ in $[1,M]$ whereby $\Y^{k}_m$ is the $m$-th component space
of bundle space $\Xk$. This component-based formulation allows us to define
empty fiber subspaces ($\Y^k_m = \Y^{k+1}_m$) and trivial fiber subspaces
($\Y^k_m = \emptyset$), which correspond to an empty set projection and an
identity projection, respectively. 

\lstset{language=C}
\alglanguage{pseudocode}

\def\Qn{Q_{new}}

\begin{figure}[ht!]
    \centering
    \algcaption{MultiRobotMotionExplorer($x^I, x^G, \X_{1:K}, \optimizer, N,
    \timebudget, \deltaS, \epsilon, \rho$)}
    \begin{algorithmic}[1]
      \State $T = \emptyset$ \Comment{Local Minima Tree}
      \While{True}
      \State $\qk = \Call{UserSelectLocalMinimum}{T}$\label{alg:explorer:selectlocalminima}

        \While{$\neg\Call{ptc}{\timebudget}$}
          \State $\Call{GrowRoadmap}{\Xkk, \qk}$\label{alg:explorer:growroadmap}
        \EndWhile
        \State $\Call{RemoveReducibleFaces}{\Skk}$ \label{alg:explorer:removereducible}
        \State $\Call{UpdateLocalMinimaTree}{T, \optimizer, N}$ \label{alg:explorer:updatetree}
      \EndWhile
    \end{algorithmic}
    \label{alg:explorer}
  \bottomrule
  \bigskip
  \algcaption{GrowRoadmap($\Xkk, \qk$)}
  \begin{algorithmic}[1]
    \State $\xr \gets \Call{ComponentRestrictionSampling}{\Xkk, \Gk, \qk, \epsilon}$\label{alg:growroadmap:samplefiber}
    \State $\Gkk \gets \Call{AddDenseGraph}{\xr, \Gkk}$\label{alg:growroadmap:dense}
    \State $\Skk \gets \Call{AddSparseGraph}{\xr, \Skk, \deltaS}$\label{alg:growroadmap:sparse}
  \end{algorithmic}
  \label{alg:growroadmap}
  \bottomrule
  
  \bigskip
\algcaption{ComponentRestrictionSampling($\Gk, \qk, \Xkk$)}
  \begin{algorithmic}[1]
    \If{$k>1$}
        \State $\x_{\Xk}^{1:M} \gets \Call{SampleBase}{\Gk, \qk, \epsilon, \rho, \Xk}$\label{alg:samplebundle:samplebase}
        \State $\xfk^{1:M} \gets \Call{Sample}{\x_{\Xk}^{1:M}, \Fk}$\label{alg:samplebundle:samplefiber}
        
        \For{each $m$ in $[1,M]$}
            \State $\xkk^m \gets \Call{Lift}{\x_{\Xk}^m, \xfk^m, \Y^m_{k+1}}$
            \label{alg:samplebundle:lift}
        \EndFor
    \Else
        \State $\xkk^{1:M} \gets \Call{Sample}{\Xkk}$ \label{alg:samplebundle:uniform}
    \EndIf
    \State \Return $\xkk^{1:M}$
  \end{algorithmic}
  \label{alg:samplebundle}
  \bottomrule
\end{figure}

Once all datastructures are initialized, we enter a while loop and ask the user
to choose a local minimum (Line \ref{alg:explorer:selectlocalminima}). In the
first iteration, we let the user automatically choose the empty set local
minimum (corresponding to the root node of the local minima tree). Once a local
minimum $\qk$ on the bundle space $\Xk$ has been selected, we then grow a
roadmap on the next bundle space $\Xkk$ while a planner terminate condition
(PTC) based on a time budget $\timebudget$ is false. The grow function is
similar to one iteration of the sparse roadmap planner \cite{dobson_2014}, with
the difference that we use a component restriction sampling procedure instead of uniform
sampling over the configuration space. 

To sample configurations we use the component restriction
sampling method, as depicted in Alg. \ref{alg:samplebundle}. We use this method to compute a biased sample on the bundle space $\Xkk$, such
that if projected onto $\Xk$, it will be close to the chosen local minimum $\qk$. If
$k$ is equal to $1$, we do not have a chosen local minimum, and we sample
uniformly on $\X_1$ by uniformly sampling each component (Line
\ref{alg:samplebundle:uniform}). If $k$ is larger than $1$, we assume a local
minimum $\qk$ on $\Xk$ is given. We then sample the base space $\Xk$ by sampling
from the graph $\Gk$ biased towards $\qk$ with bias parameter $\rho$ (Line
\ref{alg:samplebundle:samplebase}). We then perturbate the sample in an
$\epsilon$-neighborhood of $\qk$, which helps overcoming narrow passages.

Once a sample on the base space has been computed, we uniformly sample each fiber
component spaces (Line \ref{alg:samplebundle:samplefiber}). We then use the
fiber samples to lift each base space component sample into the bundle space
$\Xkk$ (Line \ref{alg:samplebundle:lift}). The lift method depends on the type
of component mapping. Currently, we support component mappings of the form
$SE(w) \rightarrow \R^w$, $\X \times \R^N \rightarrow \emptyset$ and $\X \times \R^N \rightarrow \X \times \R^M$ with
$\X = \{\emptyset, SE(w), SO(w)\}$, $w=\{2,3\}$, $0 \leq M
\leq N$, $SE$ being the special euclidean and
$SO$ the special orthogonal group, respectively.

Once the grow method terminates, we update the local minima tree (Line
\ref{alg:explorer:updatetree}). 
To update the tree, we
enumerate $N$ shortest paths on the sparse graph $\Skk$ using a depth-first
search method and let those paths converge to a local minimum using the
optimizer $\optimizer$. We then add the local minimum to the local minima tree,
if the new minimum is not straight-line deformable \cite{jaillet_2008} into an
existing minimum.

Since optimizing paths is costly, in particular for multiple robots, we do a clean up operation before updating the tree (Line \ref{alg:explorer:removereducible}). In this method we iterate over all edges in the current sparse graph. For each edge with source vertex $v_S$ and target vertex $v_T$, we compute  common neighbors $v_N$ of $v_S$ and $v_T$ and we check if the triangle $v_S, v_N, v_T$ is feasible. This operation is done by checking if the two paths $v_S$ to $v_T$ and $v_S$ to $v_N$ to $v_T$ are straight-line deformable \cite{jaillet_2008}, in which case we remove the vertex $v_N$ from the sparse graph. 

After cleaning up the sparse graph and updating the local minima tree, we return the tree and visualize it to the user.


We provide an implementation of the multi-robot motion explorer in C++ as an
extension of the Open Motion Planning Library (OMPL) \cite{sucan_2012}. We
additionally provide a graphical user interface (GUI) to visualize the local
minima tree and to let users specify fiber bundle sequences. The code is freely
available\footnote{\href{https://github.com/aorthey/MotionExplorer}{github.com/aorthey/MotionExplorer}}.

\section{Demonstrations}

To show the applicability of the multi-robot motion explorer, we demonstrate it
on a variety of multi-robot systems. We execute all demonstrations on a laptop
with a four-core $2.5$GHz processor, $8$GB Ram running Ubuntu $16.04$. We use a
minimal-length cost functional, a path optimizer provided by OMPL
\cite{sucan_2012} and we define the parameters $N=5, \delta_S = 0.15\mu,
\rho=0.05\mu$ and $\epsilon=\num{1e-3}\mu$ whereby $\mu$ is the measure of the
corresponding composite configuration space.

\subsubsection{Remark on Demonstrations}

\begin{figure}[ht]
\def\lw{\linewidth}
    \begin{subfigure}[t]{0.37\textwidth}
    \centering
        \includegraphics[width=\textwidth]{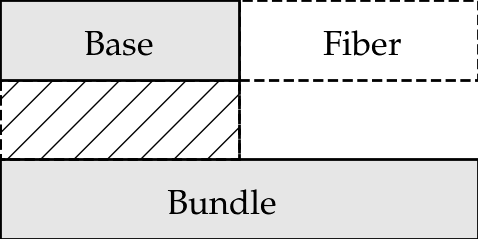}
    \caption{Template}
    \label{fig:fibers:template}
    \end{subfigure}
    \hfill
    \begin{subfigure}[t]{0.35\textwidth}
    \centering
        \includegraphics[width=\textwidth]{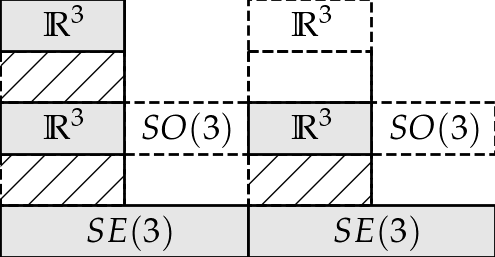}
    \caption{Two Drones}
    \label{fig:fibers:twodrones}
    \end{subfigure}
    \hfill
    \begin{subfigure}[t]{0.18\textwidth}
    \centering
        \includegraphics[width=\textwidth]{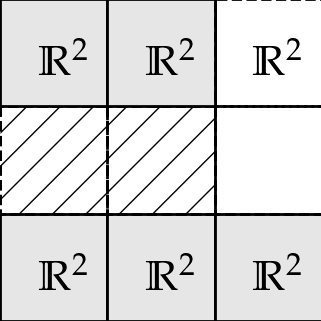}
    \caption{Three Disks (Bhattacharya Square)}
    \label{fig:fibers:3disk_bhatta}
    \end{subfigure}
    \hfill
    \begin{subfigure}[t]{0.18\textwidth}
    \centering
        \includegraphics[width=\textwidth]{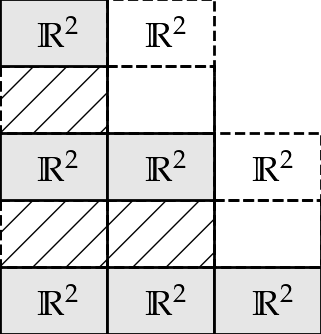}
    \caption{Three disks (Solovey Tee)}
    \label{fig:fibers:3disk_solovey}
    \end{subfigure}
    \hfill
    \begin{subfigure}[t]{0.39\textwidth}
    \centering
        \includegraphics[width=\textwidth]{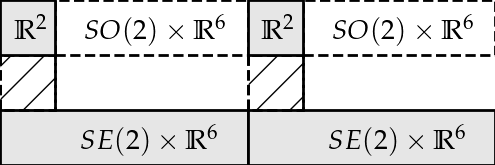}
    \caption{Two Manipulators}
    \label{fig:fibers:twomanips}
    \end{subfigure}
    \hfill
    \begin{subfigure}[t]{0.4\textwidth}
    \centering
        \includegraphics[width=\textwidth]{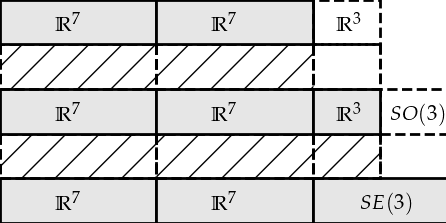}
    \caption{Two Manipulators, One Drone}
    \label{fig:fibers:twomanipsonedrone}
    \end{subfigure}    
    
    \caption{Fiber bundles reductions represented by fiber bundle diagrams. \emph{Gray rectangles}: Base or Bundle spaces, \emph{White and dashed rectangles}: Fiber spaces, \emph{Hatch patterned rectangles}: Projections from Bundle to Base space.}
    \label{fig:fibers}
\end{figure}

For each demonstration, we visualize local minima which move the robots from an
initial configuration in green to a goal configuration in red\footnote{If
printed in greyscale, initial configuration is in lightgrey, goal configuration
in darkgrey and robot during execution in white.}. Using the time budget
$\timebudget$, we run one iteration of the algorithm and report on the time and
number of local minima found. Note that the times reported are rough estimates
depending on the underlying sampling process and the chosen local minima. For
each demonstration, we chose a fiber bundle sequence, which we visualize as a diagram in Fig. \ref{fig:fibers}. Each diagram consists of the bundle space on the bottom, and the specification of the base and fiber spaces (Fig. \ref{fig:fibers:template}). We pick each fiber bundle based on faster
computation time and leading to more meaningful local minima for users of the
system. 

\subsubsection{Crossing Disks ($2$-dof)}

\begin{figure}[h]
    \centering
    \begin{subfigure}[t]{0.44\textwidth}
    \centering
        \includegraphics[width=\textwidth]{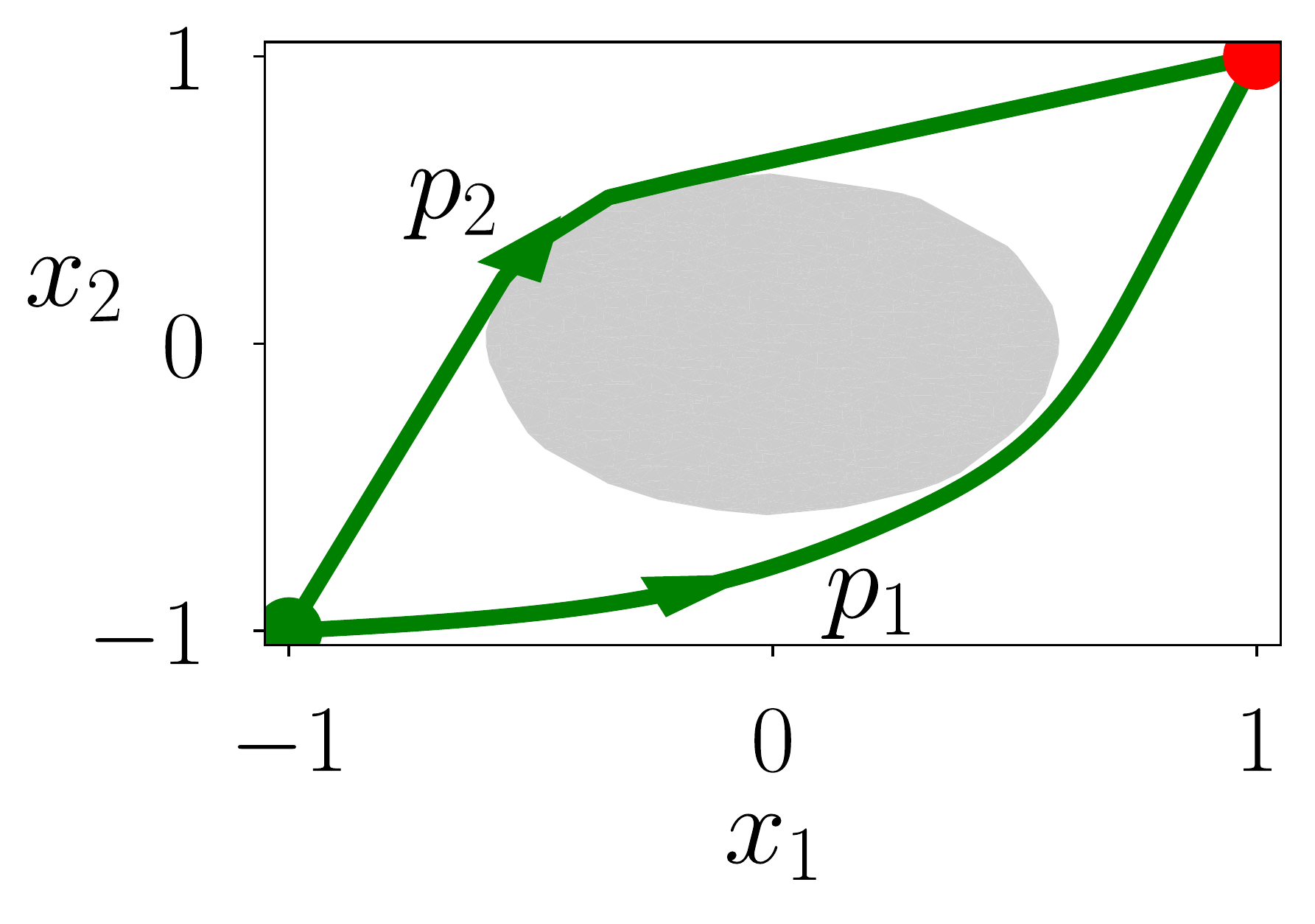}
    \caption{Configuration Space}
    \label{fig:twodisk:cspace}
    \end{subfigure}
    \begin{subfigure}[t]{0.26\textwidth}
    \centering
        \includegraphics[width=\textwidth]{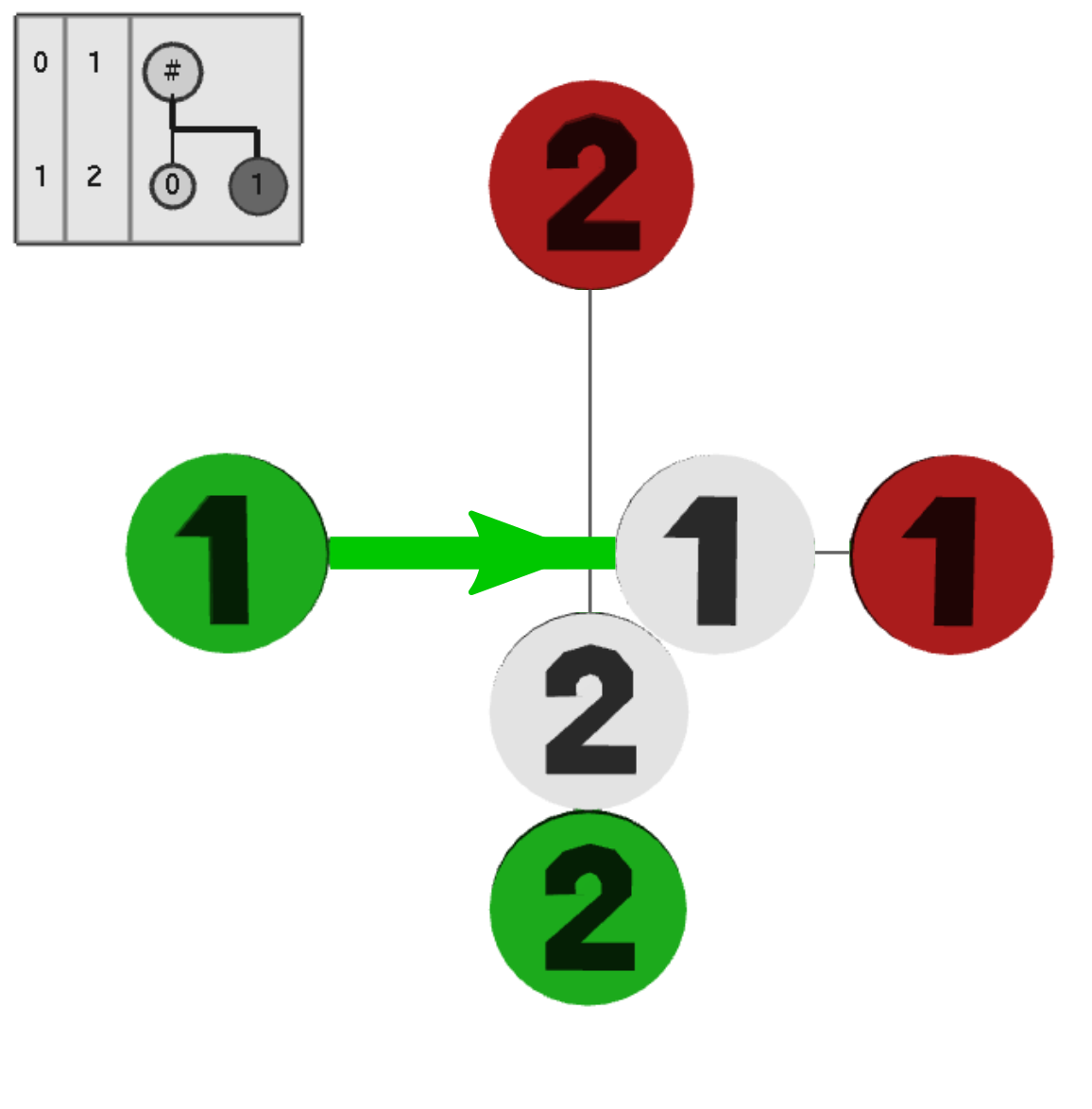}
    \caption{Minimum $p_1$}
    \label{fig:twodisk:p1}
    \end{subfigure}    
    \begin{subfigure}[t]{0.26\textwidth}
    \centering
        \includegraphics[width=\textwidth]{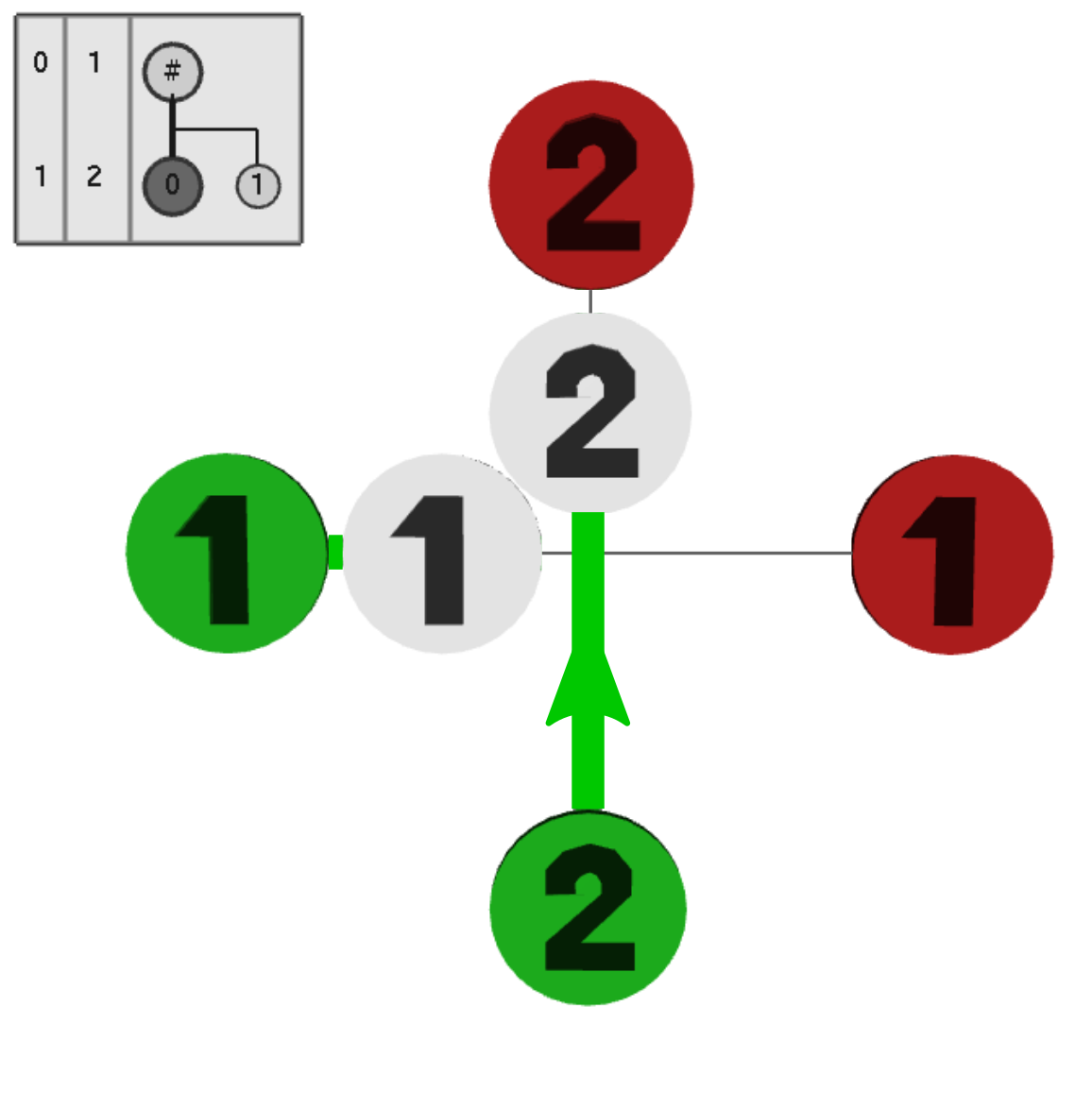}
    \caption{Minimum $p_2$}
    \label{fig:twodisk:p2}
    \end{subfigure}
    \caption{Visualizing two local minima for the crossing of two disks.}
    \label{fig:twodisk}
\end{figure}

The crossing disk problem involves two disks with equivalent radius labeled $1$
and $2$, which can each move on a line segment orthogonal to each other (Fig.
\ref{fig:twodisk}). The configuration space is a 2-d square with an infeasible circular region
caused by configurations where both disks collide (Fig.
\ref{fig:twodisk:cspace}). Using our algorithm, we find two local minima after
$0.47$s ($\timebudget=0.1$s), which we label $p_1$ and $p_2$, respectively. When
choosing minimum $p_1$, disk $1$ goes first and disk $2$ follows (Fig.
\ref{fig:twodisk:p1}). On minimum $p_2$, disk $2$ goes first and disk $1$
follows (Fig. \ref{fig:twodisk:p2}). The local minima tree is shown on the top left (see Sec. \ref{sec:foundation:localminimatree} for details).

\subsubsection{Solovey Tee ($6$-dof)\label{sec:soloveytee}}

\begin{figure}[h]
    \centering
    \begin{subfigure}[t]{0.32\textwidth}
    \centering
        \includegraphics[width=\textwidth]{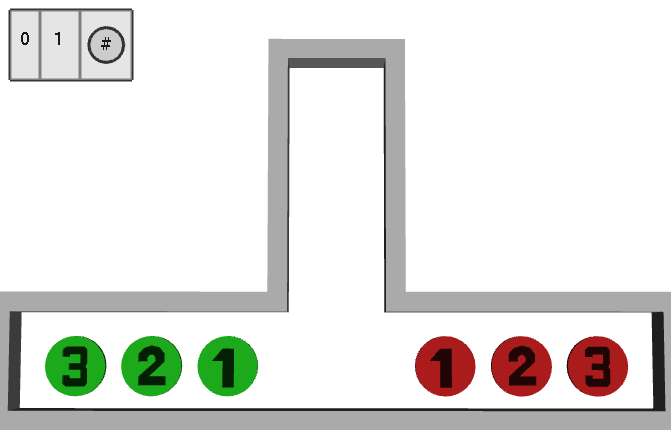}
    \caption{Initial and goal configuration}
    \label{fig:solovey:1}
    \end{subfigure}      
    \begin{subfigure}[t]{0.32\textwidth}
    \centering
        \includegraphics[width=\textwidth]{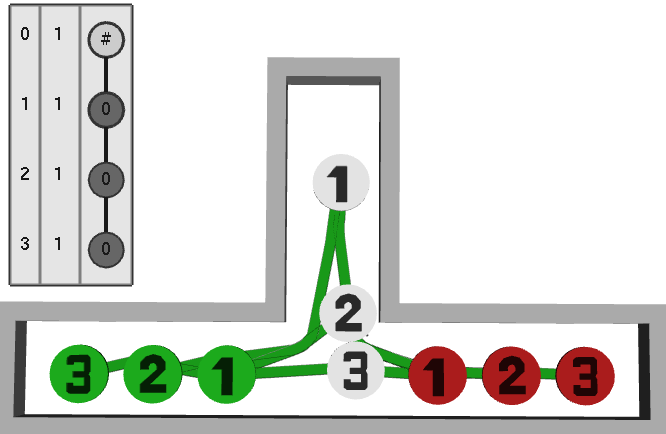}
    \caption{Local minimum using fiber bundle $(123)$}
    \label{fig:solovey:2}
    \end{subfigure}      
    \begin{subfigure}[t]{0.32\textwidth}
    \centering
        \includegraphics[width=\textwidth]{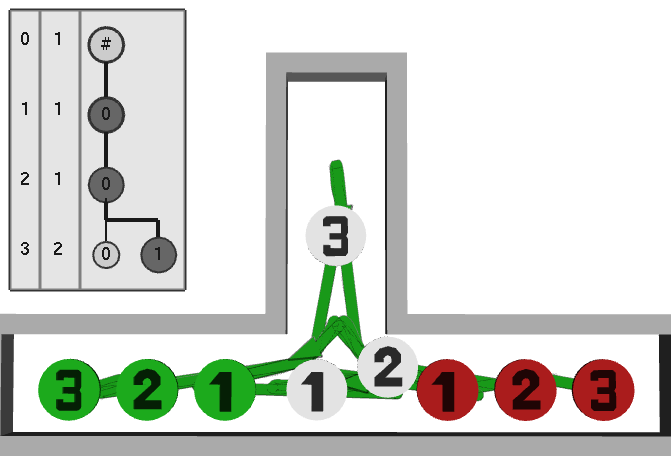}
    \caption{Local minimum using fiber bundle $(321)$}
    \label{fig:solovey:3}
    \end{subfigure}  
    \caption{Three disks in a tee (\citeauthor{solovey_2016} \cite{solovey_2016})}
    \label{fig:solovey}
\end{figure}

We next visualize local minima for three disks in a tee, a scenario proposed by \citeauthor{solovey_2016} \cite{solovey_2016} (Fig. \ref{fig:solovey:1}). We can simplify the configuration space by removing disks. We note that the sequence of removal is important for planning time. To see this, we use two different fiber bundles, one where we first remove disk $3$ and then remove disk $2$ $(321)$ and one where we first remove disk $1$ and then remove disk $2$ $(123)$. For fiber bundle $(123)$, we find one local minimum each in $0.21$s, $0.30$s and $3.07$s ($\timebudget=0.2$s), respectively (Fig. \ref{fig:solovey:2}). On the minimum, disk $3$ goes straight towards the goal, while disk $1$ and $2$ clear the path by moving into the aisle. For fiber bundle $(321)$, however, we find two local minima requiring $0.22$s, $0.43$s and $25.72$s, respectively (Fig. \ref{fig:solovey:3}). Both local minima are similar in that disks $1$ and $2$ first move towards the goal, let disk $3$ pass into the aisle, then move backwards to let disk $3$ pass towards the goal.

\subsubsection{Drones on a Tree ($12$-dof)}

\begin{figure}[h]
    \centering
    \begin{subfigure}[t]{0.32\textwidth}
    \centering
        \includegraphics[width=\textwidth]{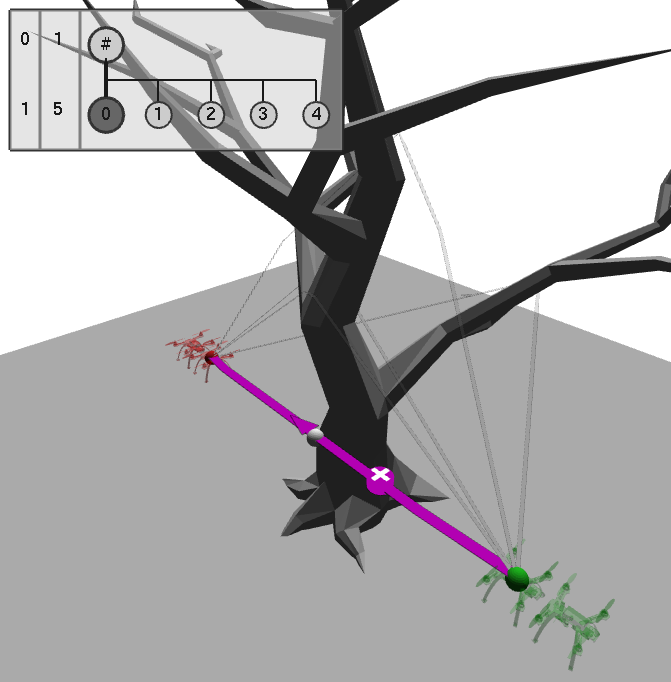}
    \caption{Minimum on first base space.}
    \label{fig:drones:1}
    \end{subfigure}  
    \begin{subfigure}[t]{0.32\textwidth}
    \centering
        \includegraphics[width=\textwidth]{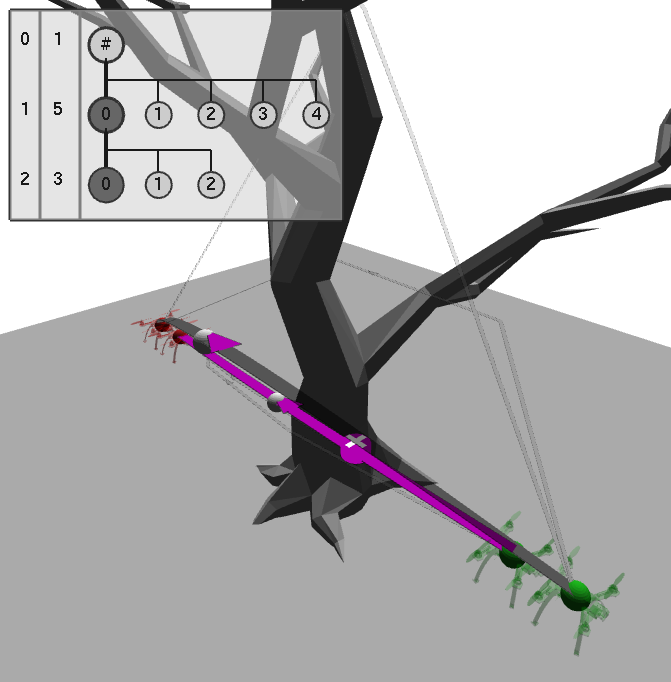}
    \caption{Minimum on second base space.}
    \label{fig:drones:2}
    \end{subfigure}  
    \begin{subfigure}[t]{0.32\textwidth}
    \centering
        \includegraphics[width=\textwidth]{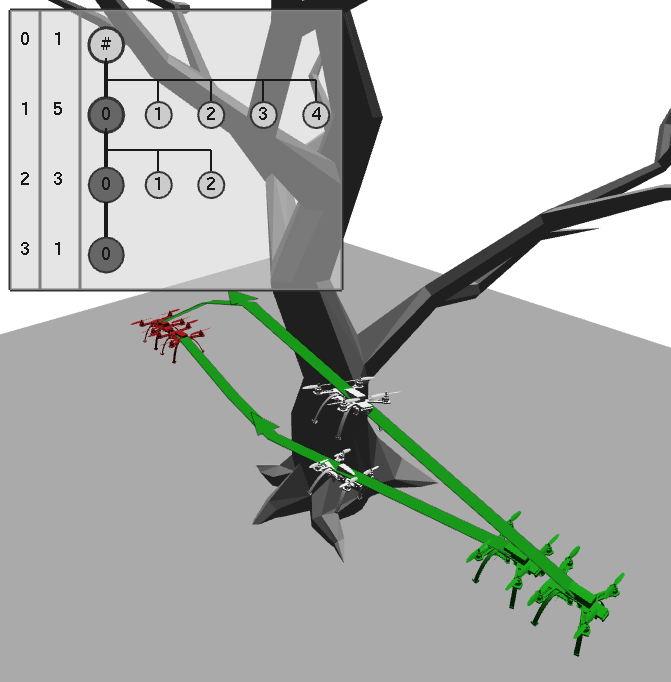}
    \caption{Local minimum path}
    \label{fig:drones:3}
    \end{subfigure}  
    \caption{Two drones flying around a tree.}
    \label{fig:dronestree}
\end{figure}

We next visualize local minima for two drones flying around a tree (Fig. \ref{fig:drones:1}). The fiber bundle reduction is given in Fig. \ref{fig:fibers:twodrones}. We find five local minima for the first drone reduction in $5.5$s ($\timebudget=0.1$s). We then select the path going left around the tree, find three minima in $2.05$s for the second drone (Fig. \ref{fig:drones:2}) and finally compute a valid local minimum in $0.19$s where both drones fly left around the tree (Fig. \ref{fig:drones:3}).

\subsubsection{Two-Arm Manipulator Baxter ($14$-dof)}

\begin{figure}[h]
    \centering
    \begin{subfigure}[t]{0.35\textwidth}
    \centering
        \includegraphics[width=\textwidth]{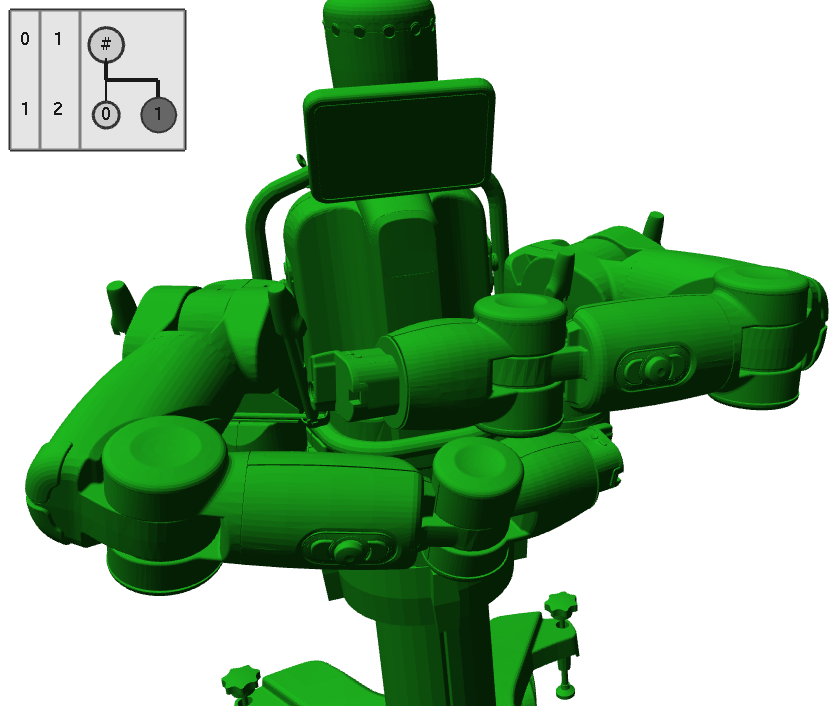}
    \caption{Start configuration}
    \label{fig:baxter:init}
    \end{subfigure}   
    \begin{subfigure}[t]{0.3\textwidth}
    \centering
        \includegraphics[width=\textwidth]{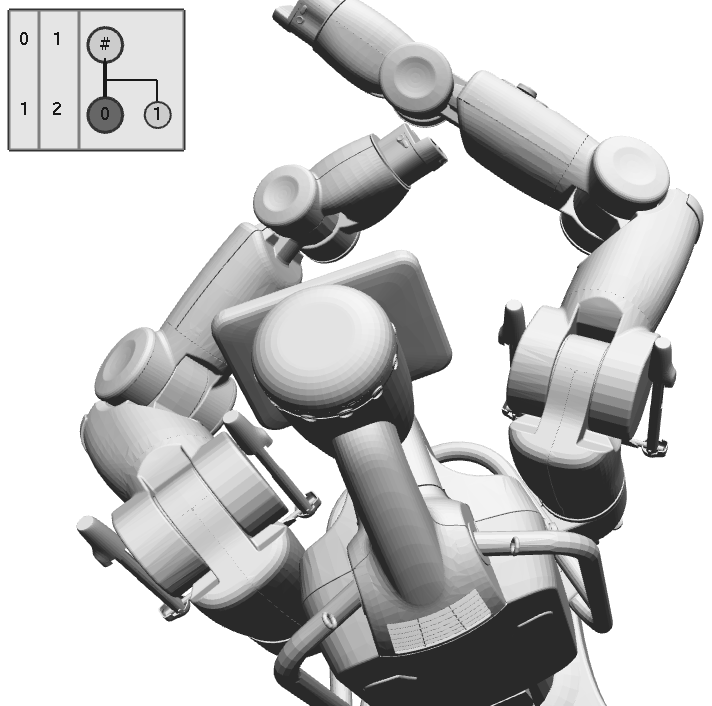}
    \caption{First minimum}
    \label{fig:baxter:p1}
    \end{subfigure}
    \begin{subfigure}[t]{0.33\textwidth}
    \centering
        \includegraphics[width=\textwidth]{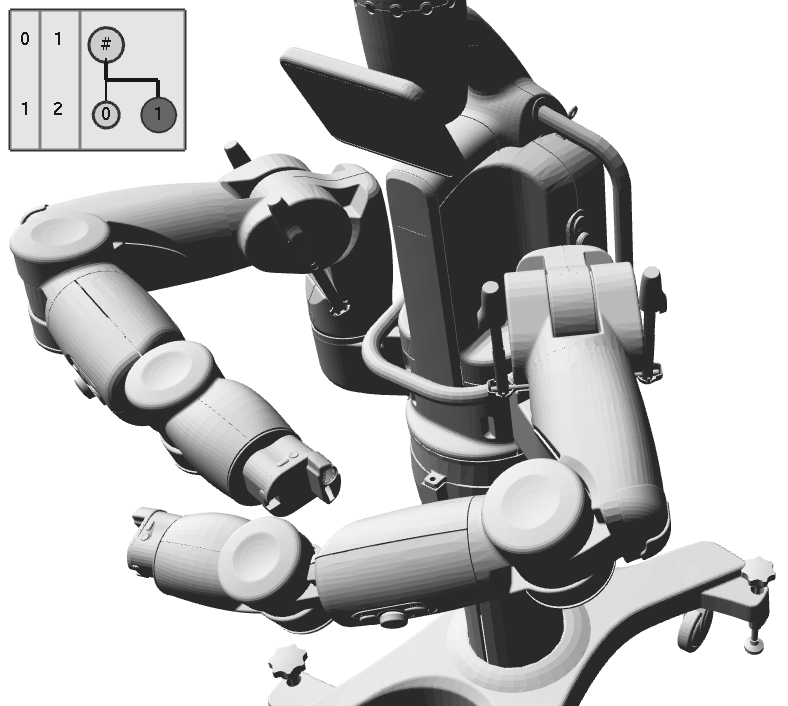}
    \caption{Second minimum}
    \label{fig:baxter:p2}
    \end{subfigure}   
    \caption{Visualizing local minima for $14$-dof baxter robot.}
    \label{fig:baxter}
\end{figure}

We next visualize local minima for the two-arm Baxter robot. We consider each arm as a separate fixed-base manipulator of $7$-dofs. The composite configuration space has $14$ dimensions. We consider a problem where Baxter has both arms in front of its torso with the left arm on top (Fig.\ref{fig:baxter:init}). The goal is to change the position of the arms, such that the right arm is on top. We find two local minima in $13.64$s ($\timebudget=10$s) planning time. On the first local minimum, the left arm is moved backward and down (Fig.\ref{fig:baxter:p1}), on the second local minimum, the left arm is moved forward and down (Fig.\ref{fig:baxter:p2}). 

\subsubsection{Manipulators Crossing ($18$-dof)}

\begin{figure}[h]
    \centering
    \begin{subfigure}[t]{0.32\textwidth}
    \centering
        \includegraphics[width=\textwidth]{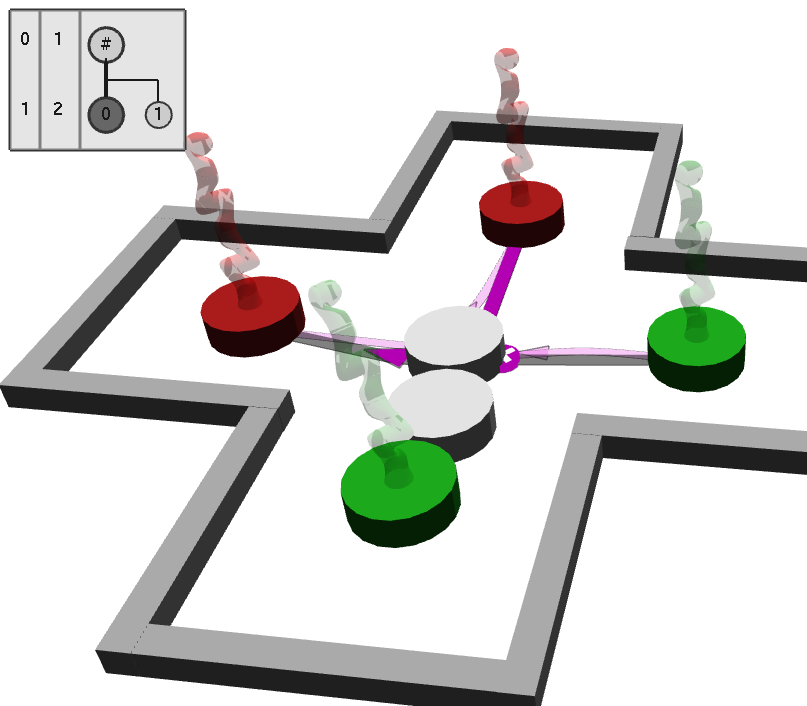}
    \caption{Reduction to two disks.}
    \label{fig:manips:1}
    \end{subfigure}  
    \begin{subfigure}[t]{0.32\textwidth}
    \centering
        \includegraphics[width=\textwidth]{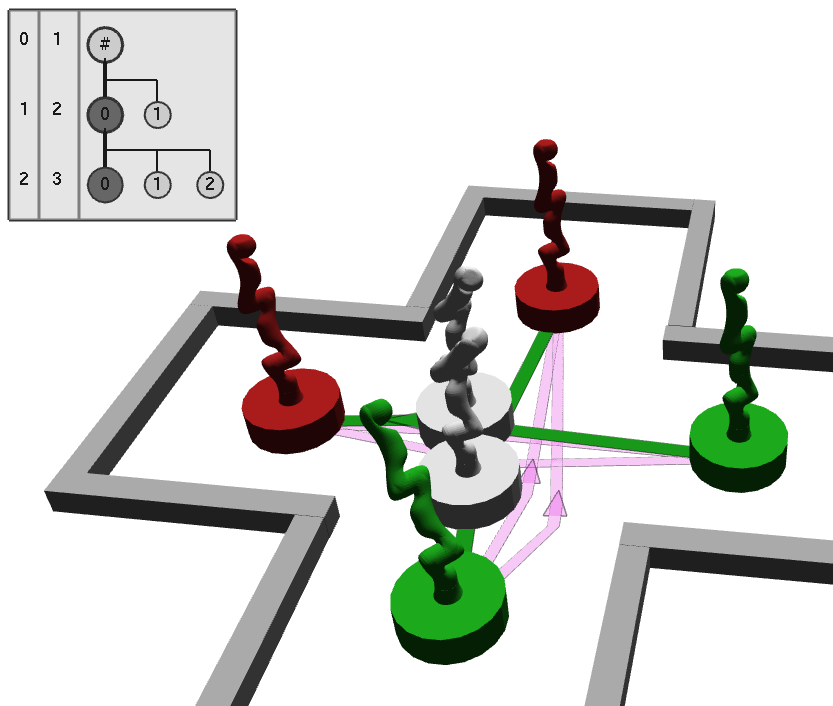}
    \caption{Local minimum $1$.}
    \label{fig:manips:2}
    \end{subfigure} 
    \begin{subfigure}[t]{0.32\textwidth}
    \centering
        \includegraphics[width=\textwidth]{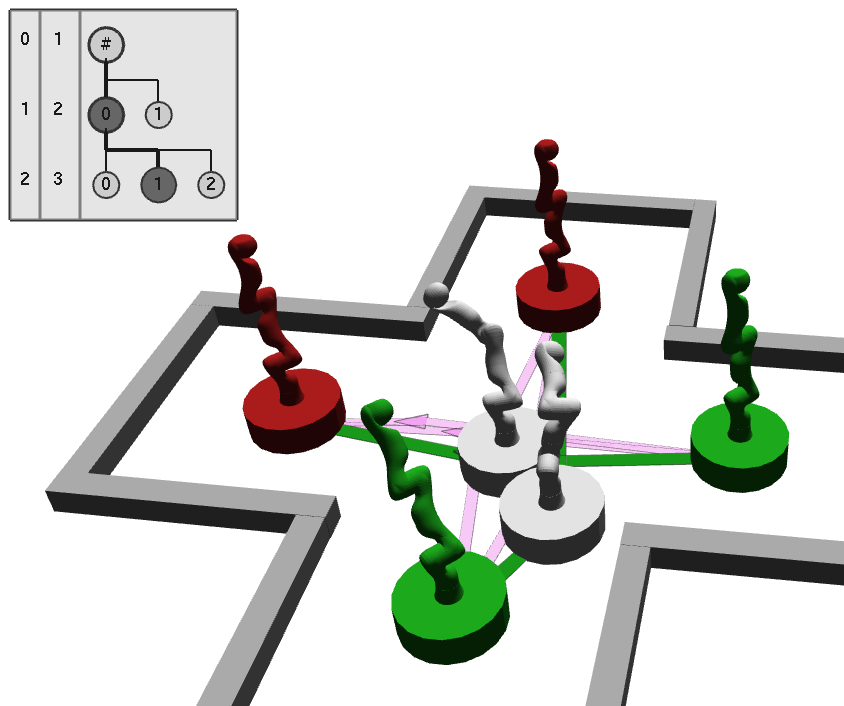}
    \caption{Local minimum $2$.}
    \label{fig:manips:3}
    \end{subfigure}
    \caption{Two manipulators navigating a crossing.}
    \label{fig:manips}
\end{figure}

We next visualize local minima for two mobile manipulators which need to cross each other to reach their goal (Fig. \ref{fig:manips}). The composite configuration space is $18$ dimensional. We use a reduction onto the base of the robots which is equivalent to two disks crossing. After planning for $0.57$s ($\timebudget=0.3$s) we find two local minima corresponding to the left manipulator going first or the right manipulator going first (Fig. \ref{fig:manips:1}). Choosing the right manipulator to go first, we then compute three local minima in $2.38$s on the composite configuration space, which correspond to different rotations of the arms (Fig. \ref{fig:manips:2} and \ref{fig:manips:3}).

\subsubsection{Drone Crossing Manipulators ($20$-dof)}

\begin{figure}[t]
    \centering
    \begin{subfigure}[t]{0.32\textwidth}
    \centering
        \includegraphics[width=\textwidth]{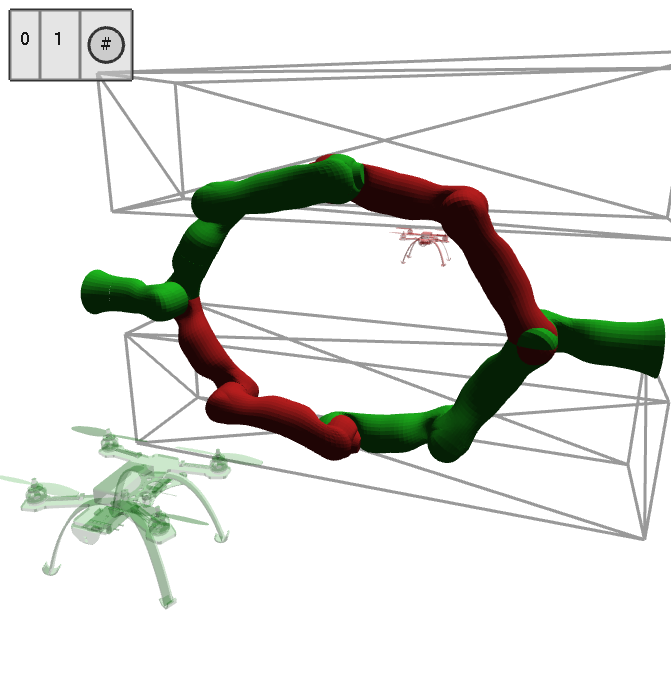}
    \caption{}
    \label{fig:manipdrone:1}
    \end{subfigure}  
    \begin{subfigure}[t]{0.32\textwidth}
    \centering
        \includegraphics[width=\textwidth]{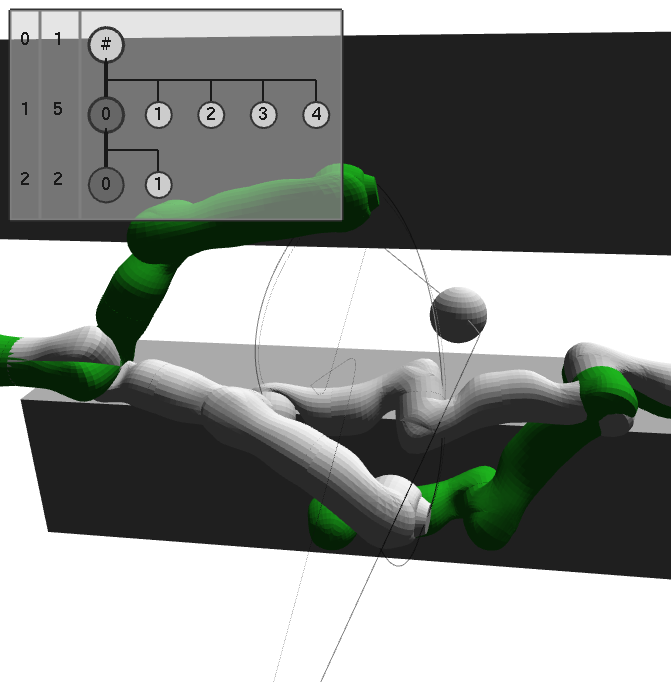}
    \caption{}
    \label{fig:manipdrone:2}
    \end{subfigure}
    \begin{subfigure}[t]{0.32\textwidth}
    \centering
        \includegraphics[width=\textwidth]{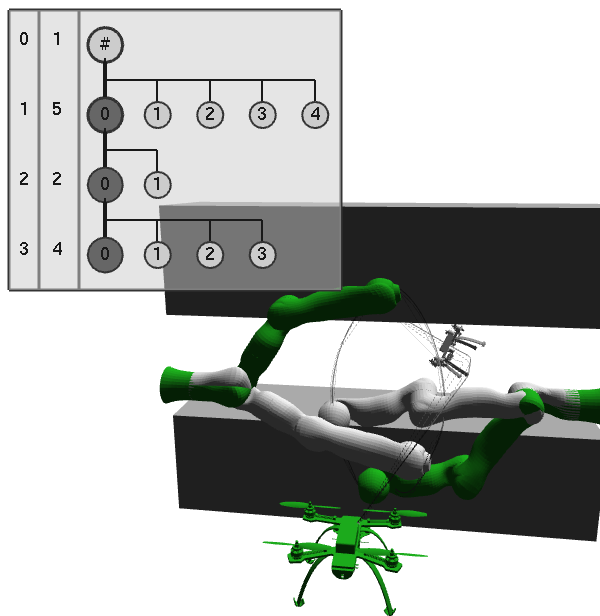}
    \caption{}
    \label{fig:manipdrone:3}
    \end{subfigure}  
    \caption{Visualizing local minima for a drone flying through two fixed-base manipulators.}
    \label{fig:manipdrone}
\end{figure}

We next visualize local minima for a drone  crossing through two fixed-base manipulator arms which have to change places (Fig. \ref{fig:manipdrone:1}). The composite configuration space has $20$ dimensions. The fiber bundle reduction is shown in Fig. \ref{fig:fibers:twomanipsonedrone}. On the lowest dimensional base space ($14$ dimensions), we compute $5$ local minima in $8.96$s ($\timebudget=1$s), whereby two minima correspond to the forward/backward motions as in the Baxer demonstration. The other minima are variations of those but with additional rotations of the joints. We then use the local minimum where the right manipulator passes behind the left manipulator to compute in $2.15$s two local minima for the inscribed sphere of the drone, one going above (Fig. \ref{fig:manipdrone:2}), one going below the right manipulator. We use the local minima going above the right manipulator to obtain four minima in $26.36$s on the bundle space (Fig. \ref{fig:manipdrone:3}). Those minima correspond to different rotations of the drone when flying above the manipulator.

\subsubsection{Bhattacharya Square ($6$-dof)\label{sec:bhattacharyasquare}}

\begin{figure}[t]
    \centering
    \begin{subfigure}[t]{0.32\textwidth}
    \centering
        \includegraphics[width=\textwidth]{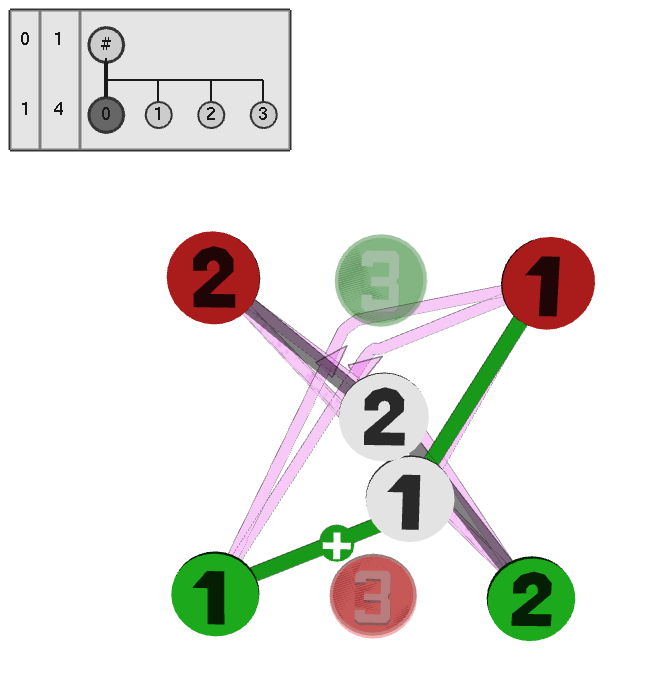}
    \caption{Minimum on Base Space}
    \label{fig:bhatta:1}
    \end{subfigure}   
    \begin{subfigure}[t]{0.32\textwidth}
    \centering
        \includegraphics[width=\textwidth]{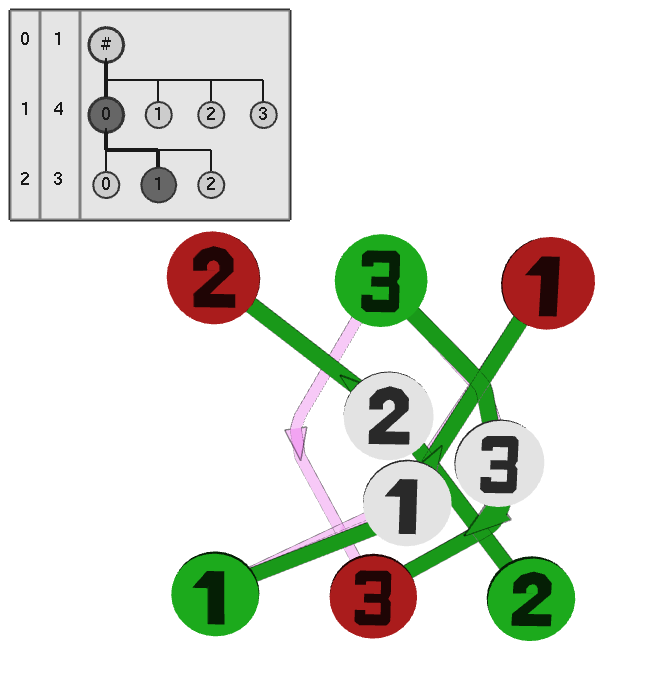}
    \caption{First minimum on Bundle Space}
    \label{fig:bhatta:2}
    \end{subfigure}   \begin{subfigure}[t]{0.32\textwidth}
    \centering
        \includegraphics[width=\textwidth]{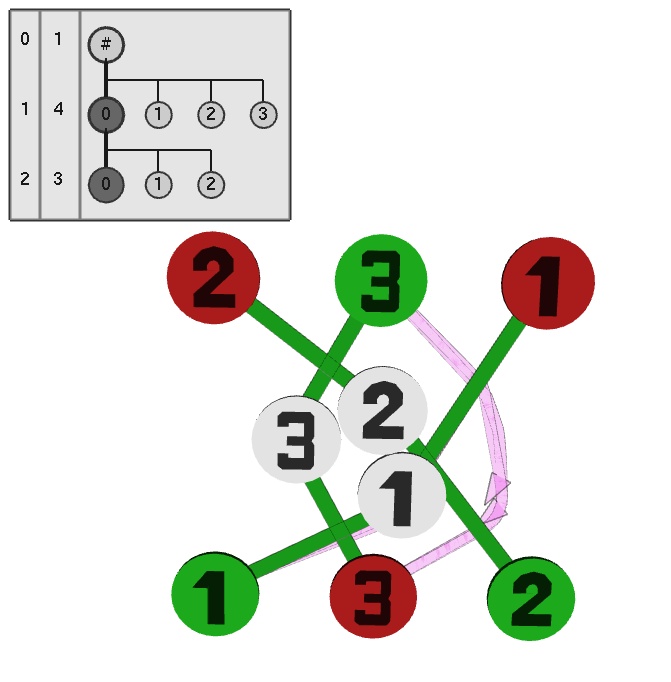}
    \caption{Second minimum on Bundle Space}
    \label{fig:bhatta:3}
    \end{subfigure}   
    \caption{Three labeled disks on a square (\citeauthor{bhattacharya_2018} \cite{bhattacharya_2018}).}
    \label{fig:bhatta}
\end{figure}

This scenario involves three labeled disks on a unit square \cite{bhattacharya_2018} (Fig. \ref{fig:bhatta}). Following the discussion by \citeauthor{bhattacharya_2018} \cite{bhattacharya_2018}, this scenario involves at least eight non-winding homotopy classes. We can obtain a meaningful grouping by removing the third disk. In that case we have two local minima depending on if disk $1$ goes first or disk $2$ goes first. Our algorithm finds both minima in $2.65$s ($\timebudget=0.3$s). As can be seen in Fig. \ref{fig:bhatta:1}, we find four local minima, two being slight variations of the desired local minima. This can be due to premature convergence of the optimizer or intricate geometric features in $4$-d space. We then select the minimum where disk $2$ goes first and find in $6.15$s three local minima on the bundle space. By inspection, we know that there should be four minima depending on if disk $3$ goes before or after disk $1$ and before or after disk $2$. However, we only find the minimum where disk $3$ goes before disk $1$ and before disk $2$ (Fig. \ref{fig:bhatta:2}) and the minimum where disk $3$ goes before disk $2$ and after disk $1$ (Fig. \ref{fig:bhatta:3}). We do not find the other local minima, most likely because they belong to narrow passages in the configuration space. We occasionally observe the algorithm to find local minima with cycling behavior, where two robots meet, cycle around each other and then continue onward. We discuss possible solutions to those problems in Sec. \ref{sec:conclusion}. 

\section{Conclusion\label{sec:conclusion}}

To visualize local minima, we developed the multi-robot motion explorer. Using
the explorer, we extended previous results on single-robot visualization
\cite{orthey_2020} by using a component-based Multilevel Morse theory framework. In demonstrations, we showed the motion explorer to
be applicable to several multi-robot scenarios involving disks, drones, and
manipulator arms.

While the algorithm works robustly on many robot platforms, we observed three
limitations.  First, we often missed local minima when the configuration space
contained narrow passages.  We could alleviate this problem by biasing sampling
towards narrow passages, by analyzing locally reachable sets \cite{sontges_2017} or by targeted sampling of undiscovered braid pattern \cite{diazmercado_2017}. Second, the algorithm can
return minima with cycling behavior, where two robots cycle around each other
before continuing. We could alleviate this problem by detecting and removing
cycles or by penalizing cycle paths using additional cost functionals.  Third,
we rely on manually specified fiber bundle reductions. To automate this, we
could specify a set of elementary planning problems and search for one which
best reduces the problem at hand.

Despite limitations, by visualizing local minima, we have contributed a useful
algorithm to the multi-robot planning toolbox. Using this algorithm, we can
increase our conceptual understanding to better debug, reduce and interact with
multi-robot motion planning problems.

\bibliographystyle{IEEEtranSN}
{\small
\bibliography{IEEEabrv, bib/general}
}
\end{document}